\documentclass{article}
\usepackage{spconf,amsmath,graphicx}
\usepackage{multirow,threeparttable, booktabs, makecell, pifont}
\usepackage{tabularx}
\usepackage{amssymb}
\usepackage{amsmath}
\usepackage{hyperref}
\usepackage{pifont}
\usepackage{url}
\usepackage[misc]{ifsym}
\usepackage{cite}
\usepackage{amsmath,amssymb,amsfonts}
\usepackage{algorithmic}
\usepackage{graphicx}
\usepackage{textcomp}
\usepackage{xcolor}
\usepackage{enumitem}
\setlist{nosep, leftmargin=14pt}

\usepackage{mwe} % to get dummy images

\title{Parameter-Efficient Fine-Tuning Medical Multimodal Large Language Models for Medical Visual Grounding}

 \name{Jinlong He$^{1 \dagger}$ \qquad Pengfei Li$^{1 \dagger}$\thanks{$^{\dagger}$Equal contribution; $^{\star}$Corresponding author} \qquad Gang Liu $^{1 \star}$ \qquad Shenjun Zhong$^{2}$}

 \address{$^{1}$ College of Computer Science and Technology, Harbin Engineering University, China \\
 	$^{2}$Monash Biomedical Imaging, Monash University, Australia
	     }
\begin{document}
%\ninept
%
\maketitle
\begin{abstract}
Multimodal Large Language Models (MLLMs) inherit the superior text understanding capabilities of LLMs and extend these capabilities to multimodal scenarios. These models achieve excellent results in the general domain of multimodal tasks. However, in the medical domain, the substantial training costs and the requirement for extensive medical data pose challenges to the development of medical MLLMs. Furthermore, due to the free-text form of answers, tasks such as visual grounding that need to produce output in a prescribed form become difficult for MLLMs. So far, there have been no medical MLLMs works in medical visual grounding area. For the medical vision grounding task, which involves identifying locations in medical images based on short text descriptions, we propose \textbf{P}arameter-efficient \textbf{F}ine-tuning medical multimodal large language models for \textbf{M}edcial \textbf{V}isual \textbf{G}rounding (PFMVG). To validate the performance of the model, we evaluate it on a public benchmark dataset for medical visual grounding, where it achieves competitive results, and significantly outperforming GPT-4v. Our code will be open sourced after peer review.
\end{abstract}
\begin{keywords}
Medical Multimodal Large Language Model, Medical Visual Grounding, Parameter-Efficient Fine-Tuning
\end{keywords}
\section{Introduction}
\label{sec:intro}

%\begin{figure*}
%	\centering
%	\includegraphics[width=0.8\textwidth]{model.eps}
%	\caption{The architecture of the model and the instruction template of the two stages.} \label{model}
%\end{figure*}

Medical visual grounding associates regions of interest (ROIs) in medical images with corresponding textual descriptions through the analysis of medical images. This is a challenging task requiring comprehensive understanding and precise alignment of visual and textual modalities. Integrating this technology into clinical workflows enhances the interpretation of medical findings within images, aiding physicians in swiftly identifying key information, accurately localizing diseases, and facilitating precise clinical decisions.

Despite its importance, medical visual grounding remains underexplored. Existing methods primarily utilize vision-language pre-training (VLP) models \cite{li2023masked,li2023self}, learning general representations by globally aligning image and text features via contrastive learning, followed by fine-tuning on medical visual grounding tasks \cite{zhang2022contrastive,wu2023medklip}. However, these approaches often lack semantic granularity, neglecting crucial local image features. To enhance semantic understanding, some studies align textual words with local image regions \cite{huang2021gloria,boecking2022making}. However, word-level alignment is constrained by contextual variability, hindering accurate capture of pathological descriptions. To overcome this limitation, Yang et al. \cite{yang2024multimodal} proposed a framework combining global and local contrastive learning, it leverages global features while precisely aligning sentence-level features with local image regions.

Recent advancements in Multimodal LLMs (MLLMs) \cite{li2023blip,he2024pefomed,gpt4}, have significantly enhanced foundational model capabilities. These models facilitate interactive image-based communication, demonstrating exceptional performance in visual tasks and complex content understanding and generation \cite{zhu2023minigpt,chen2023minigpt}. Inspired by MLLMs, we posit that an optimal solution for medical visual grounding lies within their framework. MLLMs deeply integrate cross-modal (image and text) information, improving medical knowledge representation and reasoning. They can identify subtle image details, such as lesion characteristics, and comprehend complex textual descriptions like pathology reports. In contrast, although VLP models process visual and linguistic data, their cross-modal understanding and reasoning are less profound than MLLMs, particularly with specialized medical images and terminology.

Training a medical MLLM from scratch is resource-intensive due to the need for comprehensive medical knowledge coverage, large specialized datasets, and significant computational resources. To circumvent these challenges, we propose PFMVG, a framework that leverages a foundational MLLM with Parameter-Efficient Fine-Tuning (PEFT) techniques for medical visual grounding. Specifically, we fine-tune the pre-trained weights of MiniGPT-v2 \cite{chen2023minigpt}, adapting it to medical-specific content to align medical textual and visual knowledge, followed by task-specific fine-tuning for medical visual grounding. Our resource-efficient strategy involves freezing certain model layers, such as the visual encoder, and updating critical layers like low-rank adaptation (LoRA) layers \cite{hu2021lora} for the LLM and linear projection layers for aligning image features with language model inputs. We also designed specific prompt templates for fine-tuning. Extensive experiments on the MS-CXR dataset \cite{boecking2022making} demonstrate that PFMVG achieves state-of-the-art results across eight disease categories, surpassing existing benchmarks in weighted average IoU and Dice scores. Additionally, ablation studies confirm the effectiveness of our two-stage fine-tuning process.

\begin{figure}
	\centering
	\includegraphics[width=1\columnwidth]{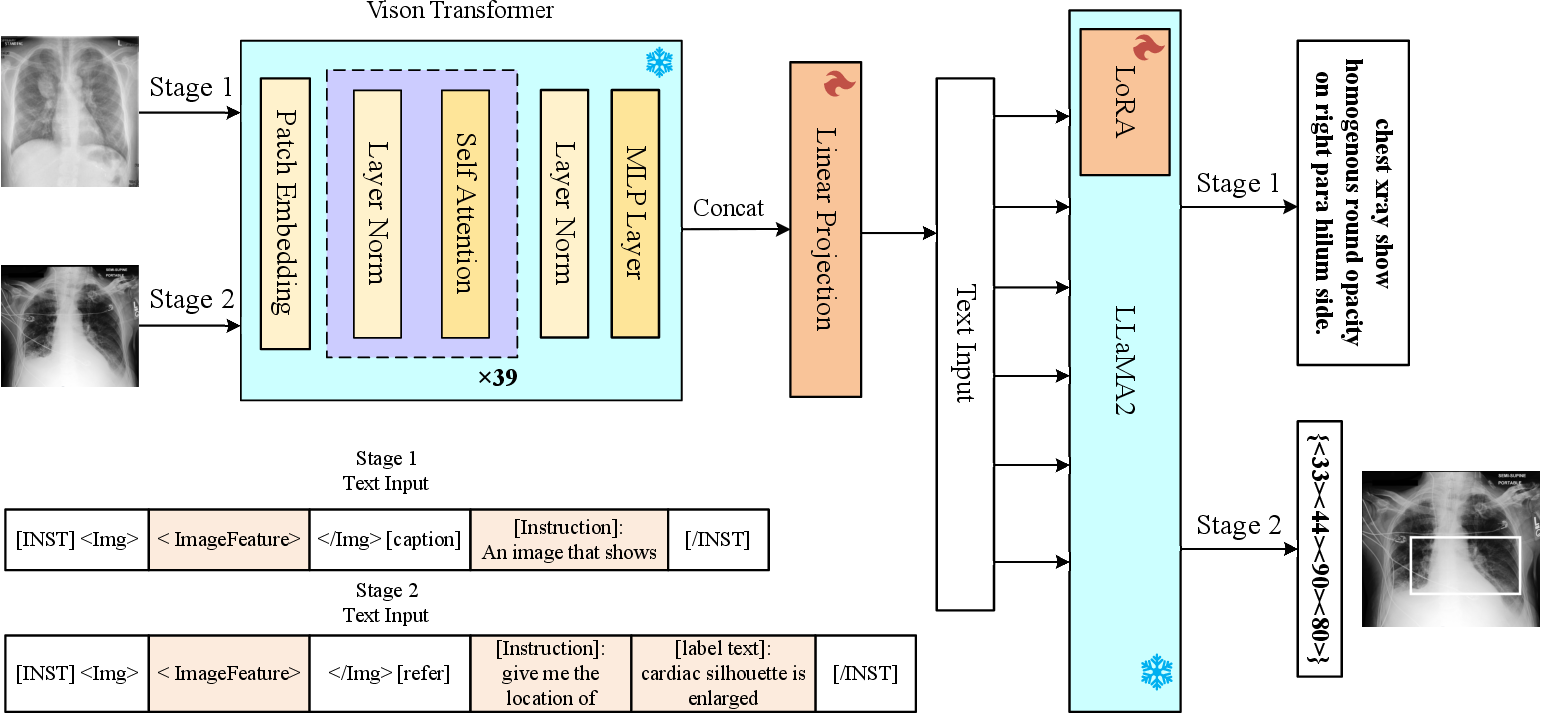}
	\caption{The architecture of the model and the instruction template of the two stages.} \label{model}

\end{figure}

\section{METHODOLOGY}
% Table generated by Excel2LaTeX from sheet 'MICCAI 2024'
\begin{table*}[htbp]
	\centering
	\caption{The IoU scores between different models evaluated on the MS-CXR dataset. wIoU represents a weighted IoU calculated taking into account the number of samples from different categories.}
	\resizebox{\textwidth}{!}{
		\begin{tabular}{c c c c c c c c c c c}
			\toprule
			\multicolumn{1}{c}{Methods} & \multicolumn{1}{c}{Pneumonia} & \multicolumn{1}{c}{Pneumothorax} & \multicolumn{1}{c}{Consolidation} & \multicolumn{1}{c}{Atelectasis} & \multicolumn{1}{c}{Edema} & \multicolumn{1}{c}{Cardiomegaly} & \multicolumn{1}{c}{\makecell{Lung \\ Opacity}} & \multicolumn{1}{c}{\makecell{Pleural \\ Effusion}} & \multicolumn{1}{c}{mean} & wIoU \\
			\midrule
			MSLL \cite{yang2024multimodal}      & 0.425     & 0.106     & 0.386     & 0.388     & 0.294     & 0.33      & \textbf{0.325} & 0.368     & \multicolumn{1}{c}{0.328} & 0.308 \\
			\multicolumn{1}{c}{MedKLIP \cite{wu2023medklip}} & \multicolumn{1}{c}{0.297} & \multicolumn{1}{c}{0.091} & \multicolumn{1}{c}{0.265} & \multicolumn{1}{c}{0.323} & \multicolumn{1}{c}{\textbf{0.327}} & \multicolumn{1}{c}{0.395} & \multicolumn{1}{c}{0.197} & \multicolumn{1}{c}{0.216} & \multicolumn{1}{c}{0.264} & 0.267 \\
			\multicolumn{1}{c}{Biovil \cite{boecking2022making}} & \multicolumn{1}{c}{0.328} & \multicolumn{1}{c}{0.137} & \multicolumn{1}{c}{0.297} & \multicolumn{1}{c}{0.275} & \multicolumn{1}{c}{0.213} & \multicolumn{1}{c}{0.406} & \multicolumn{1}{c}{0.188} & \multicolumn{1}{c}{0.224} & \multicolumn{1}{c}{0.259} & 0.281 \\
			\multicolumn{1}{c}{Gloria \cite{huang2021gloria}} & \multicolumn{1}{c}{0.29} & \multicolumn{1}{c}{0.116} & \multicolumn{1}{c}{0.304} & \multicolumn{1}{c}{0.303} & \multicolumn{1}{c}{0.201} & \multicolumn{1}{c}{0.408} & \multicolumn{1}{c}{0.197} & \multicolumn{1}{c}{0.33} & \multicolumn{1}{c}{0.269} & 0.282 \\
			\multicolumn{1}{c}{GPT-4v \cite{gpt4}} & \multicolumn{1}{c}{-} & \multicolumn{1}{c}{-} & \multicolumn{1}{c}{-} & \multicolumn{1}{c}{-} & \multicolumn{1}{c}{-} & \multicolumn{1}{c}{-} & \multicolumn{1}{c}{-} & \multicolumn{1}{c}{-} & \multicolumn{1}{c}{0.0833} & - \\
			\textbf{Ours} & \textbf{0.446} & \textbf{0.303} & \textbf{0.343} & \textbf{0.395} & 0.286     & \textbf{0.592} & 0.28      & \textbf{0.374} & \textbf{0.377} & \textbf{0.407} \\
			\bottomrule
	\end{tabular}}%
	\label{iou}%
\end{table*}%

% Table generated by Excel2LaTeX from sheet 'MICCAI 2024'
\begin{table*}[htbp]
	\centering
	\caption{The Dice scores between different models evaluated on the MS-CXR dataset. wDice represents a weighted Dice calculated taking into account the number of samples from different categories.}
	\resizebox{\textwidth}{!}{
		\begin{tabular}{c c c c c c c c c c c}
			\toprule
			\multicolumn{1}{c}{Methods} & \multicolumn{1}{c}{Pneumonia} & \multicolumn{1}{c}{Pneumothorax} & \multicolumn{1}{c}{Consolidation} & \multicolumn{1}{c}{Atelectasis} & \multicolumn{1}{c}{Edema} & \multicolumn{1}{c}{Cardiomegaly} & \multicolumn{1}{c}{\makecell{Lung \\ Opacity}} & \multicolumn{1}{c}{\makecell{Pleural \\ Effusion}} & \multicolumn{1}{c}{mean} & wDice \\
			\midrule
			MSLL \cite{yang2024multimodal}     & 0.576     & 0.163     & 0.538     & 0.538     & \textbf{0.433} & 0.485     & \textbf{0.468} & \textbf{0.525} & 0.466     & 0.44 \\
			MedKLIP \cite{wu2023medklip}   & 0.443     & 0.151     & 0.401     & 0.476     & 0.476     & 0.559     & 0.307     & 0.344     & 0.395     & 0.396 \\
			Biovil \cite{boecking2022making}    & 0.472     & 0.217     & 0.433     & 0.405     & 0.326     & 0.56      & 0.294     & 0.352     & 0.382     & 0.408 \\
			Gloria \cite{huang2021gloria}    & 0.417     & 0.181     & 0.443     & 0.442     & 0.315     & 0.567     & 0.298     & 0.476     & 0.392     & 0.407 \\
			\textbf{Ours} & \textbf{0.584} & \textbf{0.43} & 0.489     & \textbf{0.543} & 0.401     & \textbf{0.736} & 0.405     & 0.519     & \textbf{0.513} & \textbf{0.544} \\
			\bottomrule
	\end{tabular}}%
	\label{dice}%
\end{table*}%

\subsection{Model Architecture}
PFMVG features a ViT for image encoding and a LLM for image-text embeddings processing and response generation. The ViT is connected to a trainable linear projection layer. Specifically, we use the pre-trained ViT-G model from EVA \cite{eva}, which remains frozen during both fine-tuning stages. To reduce computational resources, we shorten the length of visual embeddings. By concatenating every four consecutive tokens into a single embedding, we reduce the number of visual input tokens by a factor of four. These concatenated embeddings are then fed into a trainable linear projection layer that maps them to the LLM. In this phase, the projected visual embeddings serve as the \textit{\textless ImageFeature\textgreater} component within our multimodal text input template. Combined with the question, they are input into the LLM to generate responses. We utilize the open-source, pre-trained LLaMA2-Chat (7B) as our LLM, employing LoRAfor efficient fine-tuning by updating only the LoRA parameters while keeping the rest of the model frozen.

\subsection{Two Stages Fine-tuning and Multimodel Instruction Template}
To enhance the MLLM's adherence to medical visual grounding instructions, we implement a two-stage fine-tuning process using pre-trained MiniGPT-v2 weights. Stage 1 focuses on image captioning to facilitate the MLLM's acquisition of multimodal medical knowledge. Stage 2 then advances the model's proficiency in medical visual grounding. In both stages, we employ the Llama-2 conversation template as the multimodal instruction template: \textit{[INST]\textless Img\textgreater \textless Image Feature\textgreater \textless /Img\textgreater[Task Identifier][Instruction][/INST]}. \textit{[Task Identifier]} denotes the task type, \textit{[INST]} and \textit{[/INST]} represent the user and assistant roles, respectively, and \textit{\textless ImageFeature\textgreater} contains the image embeddings encoded by the ViT.

\textbf{Stage 1 fine-tuning: medical image-text knowledge learning.} In stage 1 fine-tuning, we freeze all model parameters except the linear projection layer and the LoRA parameters. We fine-tune the model using three medical image-text datasets, ROCO \cite{pelka2018radiology}, CLEF2022 \cite{ruckert2022overview}, and MIMIC-CXR \cite{johnson1901mimic}, to enhance its understanding of medical multimodal contexts. We employ a multimodal instruction template for the image captioning task: \textit{[INST]\textless Img\textgreater \textless ImageFeature\textgreater \\ \textless /Img\textgreater [caption] [Instruction] [/INST]}.  In this template, \textit{[Task Identifier]} is replaced with \textit{[caption]}, and \textit{[Instruction]} is a randomly selected question from an instruction pool.

\textbf{Stage 2 fine-tuning: medical visual grounding.} In stage 2, similar to stage 1, we keep the ViT frozen while updating the linear projection layer and LoRA parameters. To enhance the model's proficiency in medical visual grounding, we train it on the MS-CXR dataset \cite{boecking2022making}. We employ a specialized instruction template for fine-tuning the visual grounding task: \textit{[INST]\textless Img\textgreater \textless ImageFeature\textgreater \textless /Img\textgreater [refer] [Instruction] [label text] [/INST]}. In this template, \textit{[label text]} complements \textit{[Instruction]}, the \textit{[Instruction]} is derived from an instruction pool of visual grounding questions. To improve performance and robustness, we design multiple instructions that express the same meaning. \textit{[label text]} is derived from the pathological descriptions corresponding to the medical images in the MS-CXR dataset. \textit{[Instruction]} combined with \textit{[label text]} formulates a complete question regarding the visual grounding of the relevant medical image.

During stage 2 fine-tuning, bounding boxes labeled by medical experts are converted into coordinates expressed in a specific textual format: \textit{$\{<X_{\mathrm{left}}><Y_{\mathrm{top}}><X_{\mathrm{right}}><Y_{\mathrm{bottom}}>\}$}. \textit{$<X_{\mathrm{left}}>$} and \textit{$<Y_{\mathrm{top}}>$} denote the coordinates of the bounding box's top-left corner, \textit{$<X_{\mathrm{right}}>$} and \textit{$<Y_{\mathrm{bottom}}>$} indicate the bottom-right corner's coordinates. These coordinates are normalized to the range [0, 100]. During validation, the model-generated coordinates are inversely transformed back to the original image resolution to ensure precise medical visual grounding.

\section{Experiments}
\subsection{Datasets and Implementation Details}
Four publicly available datasets were used throughout the experiments: ROCO \cite{pelka2018radiology}, MIMIC-CXR \cite{johnson1901mimic}, CLEF2022 image caption dataset \cite{ruckert2022overview}, and MIMIC-CXR dataset\cite{johnson1901mimic}. ROCO contains over 80,000 images with corresponding captions. MIMIC-CXR includes 473,057 chest X-ray images with 206,563 accompanying reports. CLEF2022 contains over 90,000 image-caption pairs. MS-CXR specifically designed for medical visual grounding. MS-CXR includes 1,153 samples with bounding boxes, covering eight diseases along with succinct radiological reports. To ensure fair evaluation, we randomly split the dataset into training, validation, and testing sets with a 7:1:2 patient ratio and assessed the model's average performance across all eight diseases.

We implemented our approach on four NVIDIA Tesla A40 GPUs. In both fine-tuning stages, we used randomly cropped $448 \times 448$ images. The MiniGPT-v2 pre-trained weights were utilized to initialize the model for the first fine-tuning stage. During the first fine-tuning stage, the model underwent pretraining for 3 epochs with a batch size of 2. This stage utilized the AdamW optimizer with a weight decay coefficient of 0.05, an initial learning rate of $1e^{-4}$, which was gradually decreased to $8e^{-5}$ following a cosine schedule. In the second fine-tuning stage, the model was trained on the MS-CXR dataset for 50 epochs with a batch size of 4. The AdamW optimizer was employed, with the learning rate decreasing from an initial $3e^{-5}$ to a final $1e^{-5}$.

% Table generated by Excel2LaTeX from sheet 'MICCAI 2024'
\begin{table*}[htbp]
	\centering
	\caption{The IoU scores from ablation study under different training setups.}
	\resizebox{\textwidth}{!}{
		\begin{tabular}{c c c c c c c c c c c c}
			\toprule
			Stage 1   & \multicolumn{1}{c}{Stage 2} & \multicolumn{1}{c}{Pneumonia} & \multicolumn{1}{c}{Pneumothorax} & \multicolumn{1}{c}{Consolidation} & \multicolumn{1}{c}{Atelectasis} & \multicolumn{1}{c}{Edema} & \multicolumn{1}{c}{Cardiomegaly} & \multicolumn{1}{c}{\makecell{Lung \\ Opacity}} & \multicolumn{1}{c}{\makecell{Pleural \\ Effusion}} & \multicolumn{1}{c}{mean} & \multicolumn{1}{c}{wIoU} \\
			\midrule
			×         & ×         & 0.118     & 0.041     & 0.104     & 0.098     & 0.108     & 0.136     & 0.137     & 0.069     & 0.101     & 0.101 \\
			×         & \checkmark         & 0.418     & 0.159     & \textbf{0.345} & \textbf{0.419} & \textbf{0.409} & \textbf{0.607} & 0.24      & 0.256     & 0.357     & 0.374 \\
			\checkmark         & ×         & 0.049     & -         & -         & -         & -         & 0         & -         & 0         & 0.006     & 0.016 \\
			\checkmark         & \checkmark         & \textbf{0.446} & \textbf{0.303} & 0.343     & 0.395     & 0.286     & 0.592     & \textbf{0.28} & \textbf{0.374} & \textbf{0.377} & \textbf{0.407} \\
			\bottomrule
	\end{tabular}}%
	\label{iouab}%
\end{table*}%

% Table generated by Excel2LaTeX from sheet 'MICCAI 2024'
\begin{table*}[htbp]
	\centering
	\caption{The Dice scores from ablation study under different training setups.}
	\resizebox{\textwidth}{!}{
		\begin{tabular}{c c c c c c c c c c c c}
			\toprule
			Stage 1   & \multicolumn{1}{c}{Stage 2} & \multicolumn{1}{c}{Pneumonia} & \multicolumn{1}{c}{Pneumothorax} & \multicolumn{1}{c}{Consolidation} & \multicolumn{1}{c}{Atelectasis} & \multicolumn{1}{c}{Edema} & \multicolumn{1}{c}{Cardiomegaly} & \multicolumn{1}{c}{\makecell{Lung \\ Opacity}} & \multicolumn{1}{c}{\makecell{Pleural \\ Effusion}} & \multicolumn{1}{c}{mean} & \multicolumn{1}{c}{wDice} \\
			\midrule
			×         & ×         & 0.184     & 0.071     & 0.167     & 0.157     & 0.181     & 0.221     & 0.212     & 0.116     & 0.164     & 0.163 \\
			×         & \checkmark         & 0.55      & 0.231     & 0.464     & \textbf{0.546} & \textbf{0.524} & \textbf{0.746} & 0.336     & 0.354     & 0.469     & 0.488 \\
			\checkmark         & ×         & 0.094     & -         & -         & -         & -         & 0         & -         & 0         & 0.012     & 0.031 \\
			\checkmark         & \checkmark         & \textbf{0.584} & \textbf{0.43} & \textbf{0.489} & 0.543     & 0.401     & 0.736     & \textbf{0.405} & \textbf{0.519} & \textbf{0.513} & \textbf{0.544} \\
			\bottomrule
	\end{tabular}}%
	\label{diceab}%
\end{table*}%

\subsection{Experimental Results}
TABLE ~\ref{iou} and ~\ref{dice} compare the performance of our model against existing models on the MS-CXR dataset, with the performance metrics for GPT-4v sourced from \cite{li2023gpt4v}, and those for other models from \cite{yang2024multimodal}. As demonstrated in TABLE ~\ref{iou}, our model surpasses the performance of existing models in six of the eight disease categories. Notably, the IoU for Pneumothorax,the disease category experiencing the most significant improvement, escalated from 0.137 to 0.303, marking an increment of 0.166. The mean IoU across the eight disease categories is 0.049 higher than that of MSLL \cite{yang2024multimodal}, representing an increase of 14.94\%. Considering the sample count in each category, the wIoU improved from 0.308 for MSLL to 0.407, an enhancement of 32.14\%. It is worth noting that the mean IoU of GPT-4v \cite{gpt4}, which is also a multimodal large model, is only 0.0833, which is much lower than the performance of our model. TABLE ~\ref{dice} presents the Dice scores performance across various models on the MS-CXR dataset.Similar to the IoU performance, the Pneumothorax category exhibits the most substantial improvement, with the Dice score rising from 0.217 to 0.43, marking an increase of 0.213.The mean Dice score across the eight categories surpasses that of MSLL by 0.047, attaining a score of 0.513. Furthermore, the wDice score sees a 23.64\% increase over MSLL's 0.44, reaching 0.544.

\begin{figure}[htb]
	\centering
	\includegraphics[width=0.7\columnwidth]{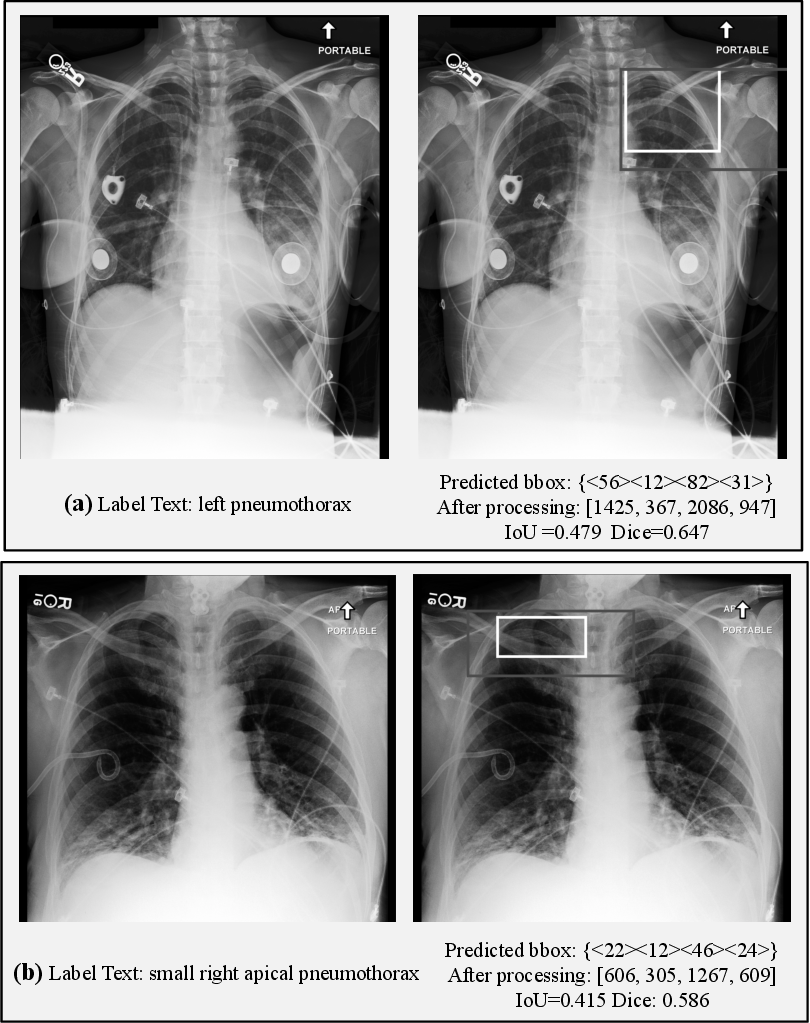}
	\caption{Results of medical visual grounding in MS-CXR. The white border is the ground-truth box, and the gray border is the box generated by the model.} \label{samples}
\end{figure}

We present the visual results of our model in Fig.~\ref{samples} It is evident that our model can locate the disease locations across various categories. As illustrated in Fig.~\ref{samples}(b), for a sample classified under "Pneumothorax", the label text is "small right apical pneumothorax". While the model-generated box does not entirely encompass the ground-truth box, it aligns more accurately with the textual description of the label.

\subsection{Ablation Study}
We conducted an ablation study to explore the impacts of the two stage fine-tuning strategy on the medical visual grounding performance. In TABLE ~\ref{iouab} and ~\ref{diceab}, they show the IoU and Dice measurements of models that (i) perform zero-shot without fine-tuning; (ii) are only fine-tuned on image caption dataset; (iii) are only fine-tuned on medical visual grounding dataset; and (iv) have two stage fine-tuning.

TABLE ~\ref{iouab} reveals that the wIoU increases from 0.101 without fine-tuning to 0.374 after only stage 2 fine-tuning, marking a significant increase of 0.273. Following exclusive fine-tuning at stage 1, the model's performance notably declines with a wIoU of merely 0.016. The model fails to generate valid outputs for five disease categories. For the remaining categories: Pneumonia, Cardiomegaly, and Pleural Effusion, the model yields valid outputs, with Pneumonia achieving an IoU of 0.049. For Cardiomegaly and Pleural Effusion, despite generating valid outputs, the IoU scores are 0. Upon completing both stages of fine-tuning, the model achieves optimal performance, with a wIoU of 0.407, marking an 8.82\% enhancement over direct stage 2 fine-tuning alone.

TABLE ~\ref{diceab} displays analogous outcomes, showing a wDice score of 0.163 without fine-tuning and an elevated wDice of 0.488 following stage 2 fine-tuning. However, following stage 1 fine-tuning exclusively, the wDice score declines significantly to 0.0311. In five disease categories, the models fail to generate valid outputs, and in two categories, while outputs are valid, the Dice scores are 0. After two-stage fine-tuning, the performance of the model reaches the best, achieving the highest Dice scores in four categories, and wDice reaches 0.544.

%\vspace{-5pt}
\section{Conclusion}
In this paper, we introduce a parameter-efficient fine-tuning medical MLLM for medical visual grounding. We adopted a two-stage fine-tuning strategy, focusing training on linear projection layers by freezing the LLM and the visual encoder. This approach not only significantly enhances the efficiency of the model training process but also minimizes resource consumption to the greatest extent, ensuring a more precise alignment with medical visual-textual knowledge. Notably, our model demonstrated superior performance on the MS-CXR dataset, significantly outperforming the GPT-4v model. Furthermore, our framework's utility extends beyond medical visual grounding tasks, showcasing its potential for a wide range of other medical multimodal tasks.

\section{ACKNOWLEDGMENTS}
This work was supported by the Ministry of Education Humanities and Social Science Research Planning Fund Project under grant number 23YJAZH079.

\section{COMPLIANCE WITH ETHICAL STANDARDS}
This research study was conducted retrospectively using human subject data made available in open access. Ethical approval was not required as confirmed by the license attached with the open-access data.
\bibliographystyle{IEEEbib}
\bibliography{refs}

\end{document}